\documentclass{IEEEtran}

\usepackage[utf8]{inputenc} 
\usepackage[T1]{fontenc}    
\usepackage{hyperref}       
\usepackage{url}            
\usepackage{booktabs}       
\usepackage{amsfonts}       
\usepackage{amsmath}       
\usepackage{nicefrac}       
\usepackage{microtype}      
\usepackage{graphicx}
\usepackage{subcaption}
\usepackage{float}
\usepackage{xcolor}
\usepackage{mathtools}
\usepackage{nomencl}
\makenomenclature

\title{Soft Attention: Does it Actually Help to Learn Social Interactions in Pedestrian Trajectory Prediction?}

\author{Laurent~Boucaud,
   Daniel~Aloise,
   and~Nicolas~Saunier,%
   \thanks{L. Boucaud was with the Department of Computer Engineering, Polytechnique Montreal (e-mail: \texttt{lrtboucaud@gmail.com})}%
   \thanks{D. Aloise is with the Department of Computer Engineering, Polytechnique Montreal (e-mail: \texttt{daniel.aloise@polyml.ca})}%
   \thanks{N. Saunier is with the Civil, Geological and Minign Engineering, Polytechnique Montreal (e-mail: \texttt{nicolas.saunier@polymtl.ca})}}

\begin{document}

\maketitle

\begin{abstract}
We consider the problem of predicting the future path of a pedestrian using its motion history and the motion history of the surrounding pedestrians, called social information. Since the seminal paper on Social-LSTM, deep-learning has become the main tool used to model the impact of social interactions on a pedestrian's motion. The demonstration that these models can learn social interactions relies on an ablative study of these models. The models are compared with and without their social interactions module on two standard metrics, the Average Displacement Error and Final Displacement Error. Yet, these complex models were recently outperformed by a simple constant-velocity approach. This questions if they actually allow to model social interactions as well as the validity of the proof. In this paper, we focus on the deep-learning models with a soft-attention mechanism for social interaction modeling and study whether they use social information at prediction time. We conduct two experiments across four state-of-the-art approaches on the ETH and UCY datasets, which were also used in previous work. First, the models are trained by replacing the social information with random noise and compared to model trained with actual social information. Second, we use a gating mechanism along with a $L_0$ penalty, allowing models to shut down their inner components. The models consistently learn to prune their soft-attention mechanism. For both experiments, neither the course of the convergence nor the prediction performance were altered. This demonstrates that the soft-attention mechanism and therefore the social information are ignored by the models.
\end{abstract}

\section{Introduction}
Predicting the trajectory of a pedestrian is a task of high importance. When it comes to self-driving cars or robots moving in crowded environments, it is key to predict the motion of surrounding pedestrians. Part of the challenge relies on understanding the \emph{social interactions}, i.e.\ the interactions between pedestrians. The path taken when walking in a crowded environment is influenced by the ones taken by the surrounding pedestrians. To predict the future path of a pedestrian by taking into account its social interactions, the authors of social-lstm~\cite{alahi2016social}, social-gan~\cite{gupta2018social} and Sophie~\cite{sadeghian2019sophie} use \emph{social information}, i.e.\ the trajectories of the other pedestrians in the scene, up to the prediction time. Given their ability to learn complex non-linear functions through a data-driven methodology, neural networks-based models have become the main method to tackle the problem of trajectory prediction using social information. In this paper they are called \emph{social models}. Amongst those, some use soft-attention~\cite{xu2015show} to better represent the social information: the learned representations of the trajectories of the surrounding pedestrians are aligned based on their relative importance in predicting the future trajectory of a pedestrian.

In the field of pedestrian trajectory prediction, approaches using soft-attention such as sophie~\cite{sadeghian2019sophie}, vain~\cite{hoshen2017vain} and social-ways~\cite{amirian2019social} are commonly assumed to use social information at prediction time. This assumption is generally based on both an ablative study of the soft-attention mechanism and the visual analysis of hand-picked predictions~\cite{gupta2018social,sadeghian2019sophie,hoshen2017vain}. Regarding the ablative studies, a model is considered to use social interactions if, compared to a baseline, it reduces the Average Displacement Error ($ADE$) and the Final Displacement error ($FDE$)~\cite{robicquet2016learning} which evaluate how close a predicted trajectory is to the ground truth.

However, several studies~\cite{nikhil2018convolutional, becker2018red} reported that on the ETH and UCY datasets, \emph{naive models}, i.e.\ models not taking social information as input, could outperform social models on both metrics. Moreover, a simple constant velocity model is shown in~\cite{scholler2019constant} to perform better than most neural networks based models, including social models. This invalidates the previous interpretations of ablative studies. Therefore, reducing, the $ADE$ and $FDE$ at prediction time cannot be considered a proof of superiority of the model using social information. This paper aims at further demonstrating, that soft-attention as currently used in the literature does not allow to take advantage of social information when predicting future trajectories.


To address this challenge, we select three soft-attention based approaches aimed at modeling social interactions: vain~\cite{hoshen2017vain}, social-ways~\cite{amirian2019social} and social-bigat~\cite{kosaraju2019social} and conduct two experiments. First, the models are trained by replacing social information with random noise and compared with the models trained with the actual social information. Second, a gating mechanism inspired from~\cite{voita2019analyzing} is introduced in the generic architecture: the output of its different components are multiplied by binary gates before being used for the final prediction. The gates are parameters learned through back-propagation by the models during training. A stochastic relaxation of the $L_0$ penalty on the parameters is used to encourage the gates to be set to zero~\cite{louizos2017learning}. In other words, the models are given the opportunity to get rid of its useless modules. 
The main contribution of this paper is the proof that soft-attention applied on social information is ignored by the models during training. 

\section{Problem definition}
In this section, the problem of trajectory prediction using social information is formally defined. To simplify the writing, we define three terms used throughout this paper. A \emph{scene} is a delimited 2D space where pedestrians move and interact, which can be an intersection or a sidewalk. The \emph{main pedestrian} refers to the pedestrian for which we currently want to predict the future trajectory. The \emph{neighbours} or \emph{surrounding pedestrians} refer to all the pedestrians in a scene at a given time, except the main pedestrian.

The problem of pedestrian trajectory prediction using social information can be formally described as predicting the next  positions of a pedestrian during $t_{pred}$ time steps (the \emph{prediction period}) based on its trajectory during the previous $t_{obs}$ time steps (\emph{the observation period}). Positions are expressed in a 2D coordinate system of the scene from a top-view perspective.

At each time $t$ there are $n$ pedestrians in the scene. We denote $x_i$ the trajectory of pedestrian $i$ during the last $t_{obs}$ time steps, expressed as:
\begin{align}
  &x_i = \{(p_i^t,v_i^t)  \mid  t \in [1,t_{obs}]\}, \qquad \bar{x}_i = \{x_j \mid \forall j \ne i,]\} \label{eq:trajectories}
\end{align}

\noindent where $\bar{x}_i$ are the trajectories of all the other pedestrians in the scene during the same period, and $p_i^t$ and $v_i^t$ are respectively the position and speed of pedestrian $i$ at time $t$. 

Our objective is to learn a function $f(. ; \theta)$ to predict $\hat{y}_i$, i.e., the future positions of pedestrian $i$, namely $\widehat{p}_i^t$, during the next $t_{pred}$ time steps.
\begin{align}
  & \hat{y}_i = f(x_i, \bar{x}_i; \theta ) =
  \{\widehat{p}_i^t  \mid  t \in [t_{obs} + 1, t_{obs} + t_{pred}]\} \label{eq:targets}
\end{align}

Given a dataset with $N$ pairs $\{(x_i, \bar{x_i}), y_i\}$, the goal is to learn the best set of parameters $\theta^*$ for the function $f(. ; \theta)$ so that the average Mean Square Error ($MSE$) between the predicted trajectory and the ground truth for all pairs is minimized:
\begin{align}
  \theta^* &=  \arg\min_{\theta} \frac{1}{N}\sum_i  MSE(y_i, \hat{y}_i; \theta)\\ 
   MSE(y_i, \hat{y}_i; \theta) &= \frac{1}{2t_{pred}} \sum_{t=t_{obs} + 1}^{t_{obs} + t_{pred}} \left\Vert p_{i}^t - \widehat{p}_{i}^t \right\Vert_2^2 \label{eq:objective}
 \end{align}
 \noindent where $y_i = \{p_i^t  \mid  t \in [t_{obs} + 1, t_{obs}+t_{pred}]\}$ is the ground truth trajectory of pedestrian $i$ and $\left\Vert . \right\Vert_2$ is the $L_2$ norm.
 The function is approximated with a neural network trained in an end-to-end fashion using the \emph{MSE} as training objective.


\section{Related Work}
The task of trajectory prediction was tackled in various ways. In their survey~\cite{lefevre2014survey}, Lefèvre~\textit{et al.} list all past works up to 2014. Kinematics models, such as the constant velocity model~\cite{ammoun2009real} or the kalman filter \cite{welch1995introduction,lin2000vehicle}, use only basic motion properties like last position, speed or acceleration to make a future prediction. Those are limited since they cannot anticipate a brutal change in motion due to external factors. Maneuver-based models consider the moving agent as executing a sequence of actions (turn-left, stop, etc.) and try to predict those actions. A clustering based approach is proposed to learn motion patterns in a scene in~\cite{morris2011trajectory}. However, models in this category do not consider social interactions. New goals for trajectory prediction are proposed in~\cite{lefevre2014survey}. Models should use their social environment and their spatial surrounding. 

Different approaches have been designed to make predictions using the social interactions and the spatial environment. Physical forces governing interactions between pedestrians are introduced as in social-forces~\cite{helbing1995social} and \cite{robicquet2016learning}. A scene is divided in cells with histograms describing the way each cell is usually traversed in \cite{ballan2016knowledge}. A Dynamic Bayesian Network is then trained to learn the transition between cells. An occupancy grid map is used in \cite{vemula2017modeling} to represent surrounding pedestrians and train a Gaussian process. The main drawback of these approaches is that they do not translate well to new scenes. To solve this issue, the focus has switched to neural networks (NN) due to their greater generalization potential. For spatial awareness, a pre-trained \emph{VGGNet} net \cite{simonyan2014very} is used to extract features from top-view images of the scenes in \cite{sadeghian2019sophie, sadeghian2018car}. Generative approaches have been designed to consider the  multi-modal nature of the trajectory prediction task~\cite{hug2017reliability}. A Conditional Variational Autoencoder

\cite{sohn2015learning} is used to generate multiple predictions~\cite{lee2017desire}. Reinforcement learning is then used to rank the predicted trajectories. Generative adversarial networks \cite{goodfellow2014generative} are used in \cite{gupta2018social,sadeghian2019sophie} to learn to predict multiple trajectories. \cite{scholler2019constant} show that those predictions do not capture the multi-modal nature of the trajectories. For social awareness, social interactions are represented through a grid-based map of their local neighbourhood and fed into a Recurrent Neural Network (RNN)~\cite{alahi2016social,gupta2018social,varshneya2017human, deo2018convolutional}. However, the authors assume that only the closest pedestrians are relevant and consider only part of them. To avoid making such assumptions, soft-attention~\cite{xu2015show} mechanisms are used in~\cite{sadeghian2019sophie,hoshen2017vain,amirian2019social,kosaraju2019social}. 

Those mechanisms are rarely evaluated in an ablative study. When they are as in~\cite{sadeghian2019sophie}, their ability to model social interactions is based on reducing the $ADE$ and $FDE$~\cite{robicquet2016learning}. Simpler NN-based models using only past motion information tend to show very good results on $ADE$, for example using a simple convolutional neural network (CNN) to encode a pedestrian past trajectory and stack a Multi-Layer Perceptron(MLP) on top to predict the future path~\cite{nikhil2018convolutional}. The same architecture is used in \cite{becker2018red}, but with a long short-term memory (LSTM) encoder instead. Surprisingly, the authors of~\cite{scholler2019constant} show that a simple constant-velocity approach could outperform more complex approaches such as those presented in~\cite{gupta2018social,sadeghian2019sophie}.  

On a final note, in the field of using deep-learning to model social interactions, most of the papers tend to follow the same working
method, which consists in taking a new successful deep learning architecture from some more mature deep learning
fields, such as computer vision or Natural Language Processing, and applying them to social interactions modeling in order to make some improvements on the traditional benchmark (ETH and UCY datasets with ADE and FDE metrics). This is a perfectly valid strategy. However, most of the time not much analysis and careful
experimentation is conducted afterwards to really understand the impact of such architectures on the modeling of
 social interactions. We don't think that improving the value of ADE and FDE on the ETH and UCY dataset is a good enough proof of the ability of a model to use social interactions. To illustrate this, in STGAT \cite{huang2019stgat} the proposed model apply Graph Attention Network to model social interactions and outperforms
 the naïve baselines on ADE and FDE. However, the trajectories are expressed in relative coordinates. It is not obvious to us that modeling social interactions from a set of trajectories not expressed in a common coordinate system would be
possible.
 Furthermore, in most studies \cite{alahi2016social, gupta2018social, sadeghian2019sophie, kosaraju2019social}, central
tendency and variation of the results are not reported. Results presented on a single run may be harder to interpret with
confidence, even more with small absolute metric improvement. For instance, in TrafficPredict \cite{ma2019trafficpredict}, the difference
 in ADE between the proposed model and a vanilla RNN is of 3.5 cm. Even though this is a 20~\% improvement we find it hard to conclude about social interactions.
 
 In this work, we focus on an attention mechanism based on soft-attention for social interactions and whether they actually make use of social information.

 

\section{The models}
In our study, we compare three approaches that include a soft-attention mechanism for social interaction modeling in their architecture: social-forces \cite{amirian2019social}, social-bigat \cite{kosaraju2019social} and vain \cite{hoshen2017vain}. We select those studies because their soft-attention mechanisms present a variety of assumptions on social interactions modeling. 

In vain, only the last observed speed and position are kept for used pedestrian. This assumes that the motion history is not critical for the task of trajectory prediction. The score function of the soft attention mechanism is not learned during back-propagation. It is basically the inverse squared Euclidean distance between the non-linear projections of the two pedestrians last  observed speeds and positions. It implicitly assumes that the Euclidean distance between pedestrians is the main factor when it comes to the importance of a social interaction.
     
In social-forces, a set of handcrafted features between the past motions of a pair of pedestrians is used to compute their attention score. There are three features: the euclidean distance, the bearing angle and the distance of closest approach \cite{pellegrini2009you}. Thus it assumes that those three features are key to determine whether a social interaction is important or not.

In social-bigat, a graph attention network is used to represent social interactions. This is very flexible and makes no assumptions on the modeling of social interactions.

\begin{figure*}[ht!]
\includegraphics[width=0.6\textwidth]{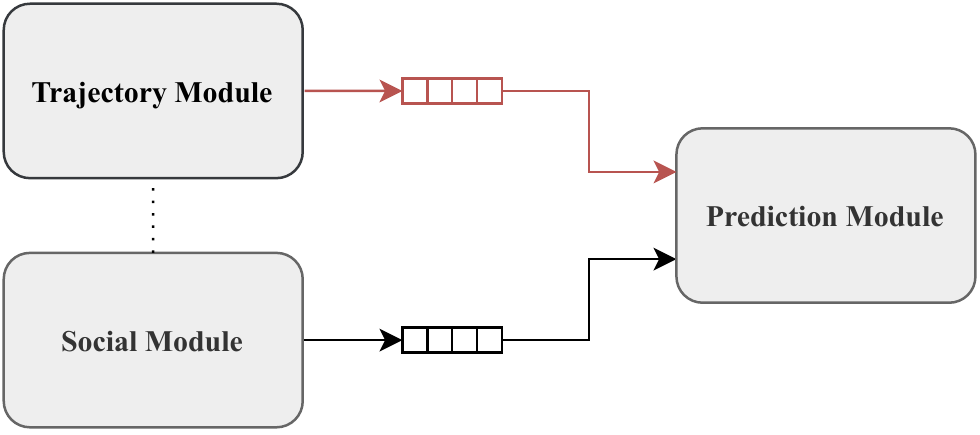}
\centering
\caption{Representation of the generic architecture. Past motion histories are fed into the trajectory module and passed to the social module (dotted line). The trajectory module output a representation of the main pedestrian's past trajectory (in red). The social module outputs a social-context tensor (in black). Both tensors are fed to the prediction module.}
\label{fig:generic}
\end{figure*}

Those three approaches, as well as most of those proposed in the literature, follow the same architecture pattern. A neural network with shared weights is used to automatically extract a features tensor from the past trajectory of each of the pedestrians in the scene, we call this part the trajectory module. A soft-attention mechanism is used to align these features tensors, forming a social-context tensor, we call this part the social module. Then the social-context tensor and the feature tensor of one pedestrian are fed into another neural network that predicts the future trajectory of this pedestrian, we call this part the prediction module. In the rest of the paper, we refer to this as the generic architecture  (\ref{fig:generic}).

We reproduced the models as described in their respective papers. However, although social-bigat and social-forces are originally trained using the framework of the Generative Adversarial Networks (GANs), we decided not to use in our study for several reasons. First, we focus on comparing soft-attention mechanisms, therefore we need to choose only one training framework to allow for a  fair comparison between the three models. We choose not to use GANs because of the way they are usually evaluated in this field. At prediction time, $k$ trajectories are generated and the best one is selected using the groundtruth. The ADE and FDE are reported on the best selected trajectory only. Using the groundtruth trajectory to select the best predicted trajectory is not a correct practice and we therefore prefer to avoid it. Furthermore, \cite{scholler2019constant} pointed out that it is similar as generating a set of constant velocity trajectories with a random angle and selecting the best one. The models are trained using the Mean Squared Error between the predicted and groundtruth trajectories as loss function through backpropagation as in social-lstm~\cite{alahi2016social}, vain~\cite{hoshen2017vain} or cidnn~\cite{xu2018cidnn}.

\section{Do they use social information?}
In this section, we study in two experiments whether the prediction module uses the output of the social module when making its prediction. 

\begin{figure*}[ht!]
\includegraphics[width=0.6\textwidth]{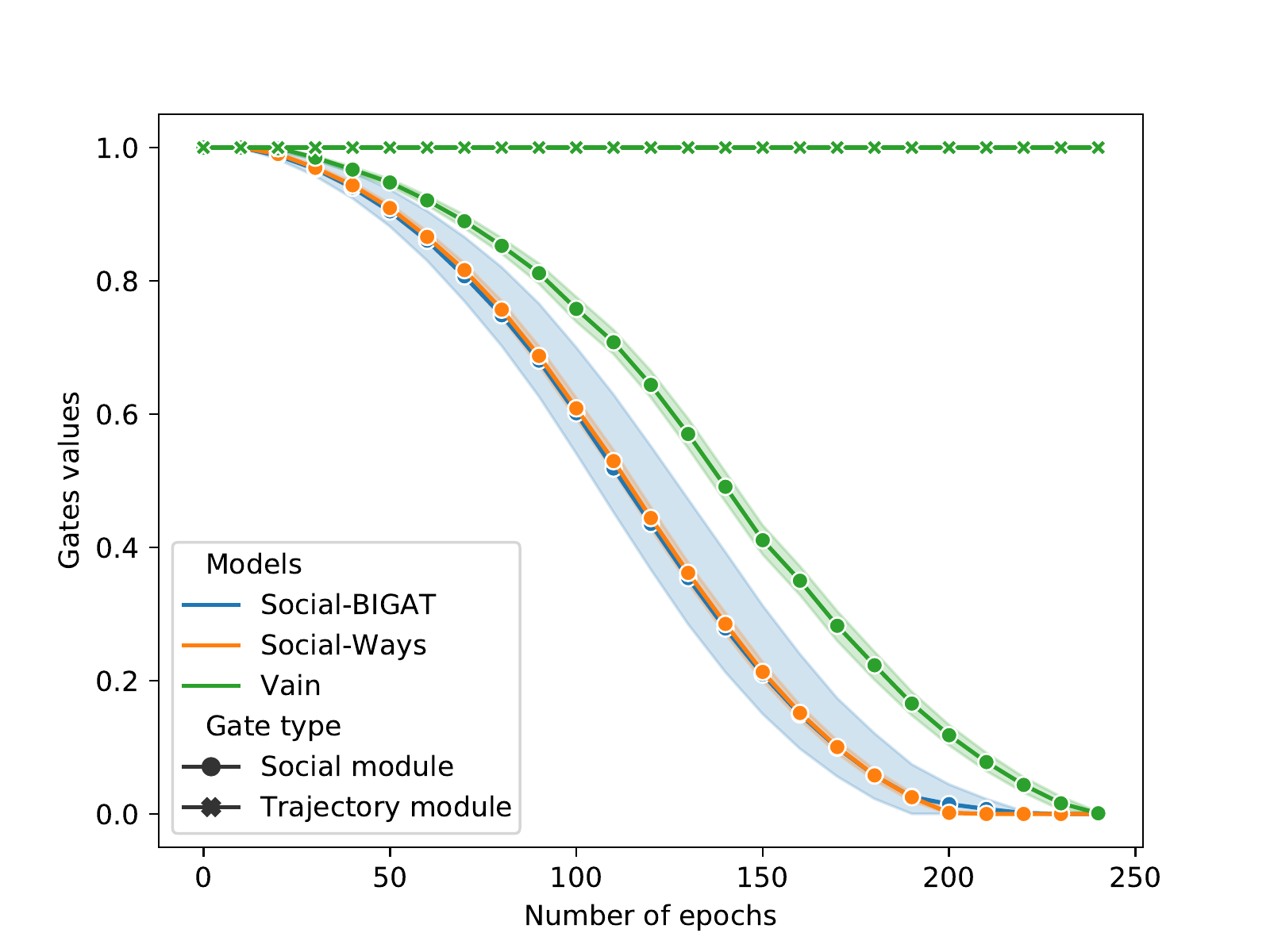}
\centering
\caption{Illustration of the gating mechanism on a simplified version of the generic architecture. The feature vector $\widetilde{T}_i$ of the main pedestrian $i$ is obtained by multiplying the output of the trajectory module $T_{i}$ with the binary gate $g_\tau$. Similarly, the social context vector $\widetilde{A}_i$ is obtained by multiplying the output of the social module $A_{i}$ with the binary gate $g_a$. They are then fed into the prediction module.}
\label{fig:gates}
\end{figure*}


\subsection{Random noise as social information} \label{sec:random} 
In the first experiment, the social information is replaced with random noise when training the models. Specifically, at training time for a pedestrian, we generate a random number of neighbours $r$ from a uniform distribution between zero and the maximum number of agents encountered in the dataset and then randomly generate $r$ random trajectories. A random trajectory is generated by randomly picking $2 \times t_{obs}$ values from a uniform distribution between 0 and 1. The generated points all belong in the space of the positions normalised across all the scenes (see section~\ref{sec:methodo}). These random trajectories are then fed into the social module. Once the training is done, we observe whether training with random social information has an impact on the convergence and performance of the models. 
\subsection{Gates and $L_0$ regularization} \label{sec:gates}
In the second experiment, we study whether the soft attention mechanism helps minimizing the training objective defined in equation~\ref{eq:objective}. We argue that the best way to find out is to give the models the opportunity to shut down some of the information flow while training. To that end we use a gating mechanism inspired from~\cite{velivckovic2017graph}. In this work, they apply binary gates on the outputs of the heads of a transformer architecture. The authors verify how many heads they can prune without harming the model prediction. Similarly, we define two binary gates as depicted in Figure~\ref{fig:gates}. Those gates are parameters learned during the training phase. We want them to take values of either 0 or 1. Based on the work of~\cite{louizos2017learning}, 
we first reparametrize the outputs vectors of the \emph{trajectory and social modules} using two binary gates, respectively the trajectory gate $g_\tau$ and the attention gate $g_a$ as follows:
\begin{equation}
   \widetilde{A}_i = A_{i}g_a, \quad \widetilde{T}_i = T_{i}g_\tau, \quad    g_a, g_\tau \in \{0,1\}, \quad \boldsymbol{g} = \begin{bmatrix} g_{\tau} \\  g_{a} \end{bmatrix}  \label{eq:reparam} 
\end{equation}
Since the gates are learned, the network has the possibility to set them to zero, thus ignoring either the information from the social module or from the trajectory module, or both. To encourage such behavior, we add a $L_0$ norm penalty term to the cost function when training. The $L_0$ norm is expressed as the number of non zero gates (equation~\ref{eq:l0disc}). 
\begin{equation}
   L_0(\boldsymbol{g}) = g_a + g_\tau   \label{eq:l0disc} 
\end{equation}
However, the $L_0$ norm is not differentiable, and therefore, we cannot train the models through back-propagation using the penalty term as it is. To solve this issue, a stochastic relaxation of the $L_0$ norm was designed in~\cite{louizos2017learning}, where gates are  sampled from a continuous distribution over $[0,1]$. The distribution is chosen specifically to allow the reparametrization trick~\cite{kingma2013auto}. It can be expressed as $\boldsymbol{g} \sim H( \boldsymbol{\phi}, \boldsymbol{\epsilon})$ where $\boldsymbol{\epsilon}$ are the parameters of a free noise distribution and $\boldsymbol{\phi}$ are the parameters of a deterministic transformation. The $L_0$ norm is then approximated by the sum over the gates of their probability of being non-zero as follows:

\begin{equation}
      L_0(\boldsymbol{g};\boldsymbol{\phi}) = \sum_{k \in \{a,\tau\}} (1 - P(g_k = 0| \phi_k)) \label{eq:l0}
\end{equation}

\noindent  where $\phi_k$ is the distribution parameter for the sampling of gate $k$.

The training objective with the $L_0$ penalty term is expressed as:
\begin{equation} L(\hat{y}_i,y_i; \boldsymbol{\theta},\boldsymbol{\phi}) = MSE(\hat{y}_i,y_i; \boldsymbol{\theta}) + \lambda L_0(\boldsymbol{g};\boldsymbol{\phi}) \label{eq:objective2}
\end{equation}
\noindent  where $\lambda$, the strength of the penalty, is a hyper-parameter. 

Thanks to this reformulation of the gating mechanism, the models can be optimized through back-propagation using a Monte-Carlo approximation. In practice, when training models using batch gradient descent, we re-sample $\boldsymbol{g}$ for every batch. At test time, gates values are estimated using a deterministic expression given in \cite{louizos2017learning}. As in the original paper \cite{louizos2017learning}, we use the Hard-Concrete distribution, with parameters $\boldsymbol{\phi} = (\boldsymbol{\alpha}, \beta)$. A fixed value of 0.5 is used for $\beta$ and $\boldsymbol{\alpha} = \begin{bmatrix} \alpha_{\tau} \\  \alpha_{a} \end{bmatrix}$ are the parameters learned through back-propagation to change the gate values at test time. There is one value per gate. 

In practice, the distribution parameters $\boldsymbol{\alpha}$ are first initialized so that the deterministic evaluations of gate values are one. The gates are frozen and the models are trained without the $L_0$ penalty. This results in regular training of the models. Once the models converge, we unfreeze the gates and add the $L_0$ penalty term to the training objective as in equation~\ref{eq:l0}. The models are thus learnt from both the motion history of the main pedestrian and the social information. Then when training is done, the models are given the opportunity to get rid of their modules by setting their respective gates to zero. 

Our research hypothesis is that the attention gate will be put to zero for all models. Furthermore, the prediction performance is expected to be maintained or even improved (removing some noise, training for longer) when the attention gate is set to zero. If this is the case, then applying a soft-attention mechanism on the social information is useless in training a NN for the task of trajectory prediction described in this paper.
\section{Methodology} \label{sec:methodo}

\subsection{Baselines}
Three baseline models that do not use social information are studied along with the four social models. First baseline is the \emph{Constant velocity(CV)} presented in~\cite{scholler2019constant}. The  second baseline from~\cite{becker2018red} is an MLP fed with features extracted from the motion history of a pedestrian using a LSTM to predict the pedestrian's future path, referred to as \emph{LSTM-MLP}. The third baseline is an MLP that takes as input the last speed and position of a pedestrian to predict its future motion, referred to as \emph{BASIC-MLP}.

\subsection{Implementation details} \label{ssec:1}
\textbf{Datasets}: We use the standard benchmark with the two datasets \textsf{ETH} \cite{pellegrini2009you} and \textsf{UCY}~\cite{lerner2007crowds}. Trajectories are sampled every 0.4~s. For each pedestrian, we observe eight time steps and predict the following twelve. Sub-trajectories of 20~positions are extracted. The trajectory of a neighbour is included as social information if it is in the scene during the last observation time step. We use zero padding for incomplete trajectories and min-max scaling on the positions. Data augmentation is performed by randomly rotating the trajectories.

\textbf{Evaluation}: Trajectories are recorded across five scenes. Each model is trained on four scenes and evaluated on the remaining one. For each scene we get the mean $ADE$ and mean $FDE$.  The test $ADE$ and $FDE$ are the average $ADE$ and $FDE$ across scenes. For hyper-parameter optimization, the data on four scenes is split in a training set and a validation set. Models are implemented using \emph{Pytorch} and trained using a \textsf{NVIDIA GTX960m} GPU. The code will be made available on Github. Every experiment is repeated five times and mean and standard deviation are reported.

\textbf{Training}: The models were trained through back-propagation using the Adam optimizer, a learning rate of 0.001 and a batch size of 32. Social models are trained for around 20 epochs whereas naive models are trained for around 10 epochs. All models converged and none of the learning curves showed signs of overfitting.

\textbf{Regularization}: $\lambda$, the strength of the $L_0$ penalty, is set to a value of 0.005. Therefore, the maximum value of the $L_0$ penalty is 0.01 if both gates are equal to one.

\subsection{Metrics}
Two metrics are used as in the standard benchmark designed by \cite{robicquet2016learning}. The $ADE$ and $FDE$ between a predicted trajectory $\hat{y}_i$ and the ground truth $y_i$ are expressed as: 
\begin{align}
    ADE(y_i, \hat{y}_i) & = \frac{1}{t_{pred}} \sum_{t=t_{obs + 1}}^{t_{end}} \left\Vert p_{i}^t - \widehat{p}_{i}^t \right\Vert_2\\
    FDE(y_i,\widehat{y}_i) & = \left\Vert p_{i}^{t_{end}} - \widehat{p}_{i}^{t_{end}} \right\Vert_2 \label{eq:metrics}
\end{align}
\noindent where $t_{end} = t_{obs} + t_{pred}$. 

\section{Results}

\paragraph{Regular training} The models are first trained regularly and their $ADE$ and $FDE$ are compared. The results can be found in Table~\ref{fig:ade_fde}. The reported mean \emph{ADEs} range between 0.56~m and and 0.7~m whereas the average $FDE$ range between 1.19~m and 1.4~m. As reported in \cite{scholler2019constant}, \emph{CV} acts as a performance lower bound. As reported in \cite{becker2018red}, the naive models outperform the social models. It seems that the simpler the model (fewer degrees of freedom), the better the $ADE$ and $FDE$.

\begin{table*}[h!]
\caption{ADE and FDE in meters for regular training}
\label{fig:ade_fde}
\centering
\begin{tabular}{l|l|l|l|l|l|l|l}
\toprule
    &  &      Basic-MLP &       LSTM-MLP &   Social-BIGAT &    Social-Ways &           Vain & CV \\
\midrule

    & eth &  \textbf{1.07 +/- 0.02} &  1.08 +/- 0.03 &  1.11 +/- 0.04 &  1.08 +/- 0.02 &  \textbf{1.07 +/- 0.02} & 1.09\\
    & hotel &  0.42 +/- 0.06 &  0.41 +/- 0.04 &  0.48 +/- 0.08 &   0.5 +/- 0.01 &  0.48 +/- 0.02& \textbf{0.35} \\
 ADE   & univ &  0.57 +/- 0.01 &  0.71 +/- 0.02 &  0.79 +/- 0.04 &  0.76 +/- 0.03 &   0.7 +/- 0.03& \textbf{0.54}\\
    & zara1 &  0.48 +/- 0.01 &  0.53 +/- 0.01 &  0.56 +/- 0.02 &   0.5 +/- 0.01 &  0.47 +/- 0.01& \textbf{0.46}\\
    & zara2 &  0.36 +/- 0.02 &  0.46 +/- 0.01 &   0.5 +/- 0.03 &  0.45 +/- 0.01 &  0.45 +/- 0.02 & \textbf{0.34}\\
    \hline
 AVG    &  &  0.58 +/- 0.02 &  0.64 +/- 0.01 &  0.69 +/- 0.02 &  0.66 +/- 0.01 &  0.63 +/- 0.01& \textbf{0.56} \\
     \hline\hline

    & eth &  2.14 +/- 0.04 &  2.17 +/- 0.05 &  2.18 +/- 0.05 &  2.11 +/- 0.03 &   \textbf{2.1 +/- 0.03} & 2.32\\
    & hotel &   0.8 +/- 0.11 &  0.76 +/- 0.12 &   0.9 +/- 0.16 &  0.94 +/- 0.04 &   0.9 +/- 0.05 &\textbf{0.67}\\
 FDE   & univ &  1.24 +/- 0.03 &  1.42 +/- 0.05 &  1.57 +/- 0.09 &  1.47 +/- 0.01 &  1.43 +/- 0.07 & \textbf{1.16}\\
    & zara1 &  1.07 +/- 0.02 &   1.1 +/- 0.02 &  1.14 +/- 0.04 &  1.01 +/- 0.02 &  \textbf{0.96 +/- 0.02} & 1.03\\
    & zara2 &  0.78 +/- 0.03 &  0.91 +/- 0.01 &  0.98 +/- 0.05 &  0.89 +/- 0.03 &  0.88 +/- 0.04 & \textbf{0.76}\\
    
    \hline
 AVG    &  &  1.21 +/- 0.03 &  1.27 +/- 0.02 &  1.35 +/- 0.04 &  1.29 +/- 0.01 &  1.25 +/- 0.02& \textbf{1.19} \\
\bottomrule
\end{tabular}
\end{table*}

\paragraph{Random training} The social models are then trained with randomly generated neighbour trajectories as described in section~\ref{sec:random}. We don't observe any changes in the models' convergences. The resulting ADE and FDE are reported in Table~\ref{fig:exp_bigat} for Social-Bigat under the column Social-Bigat(random) , in Table~\ref{fig:exp_ways} for Social-Ways under the column Social-Ways(random) and in  Table~\ref{fig:exp_vain} for Vain under the column Vain(random). Models trained with random information all exhibit slightly better $ADE$ and $FDE$ than their regularly trained counterparts. However, the difference is lower or equal than 0.05~m which is negligible for the task of trajectory prediction. Therefore, replacing the social information with random noise did not have a significant influence on the accuracy of the predicted trajectories.

\begin{table*}
\caption{ADE and FDE in meters for Social-Bigat on the three experiments(normal, random and gates)}
\label{fig:exp_bigat}
\centering
\begin{tabular}{l|l|l|l|l}
\toprule
    &       & Social-BIGAT(normal) & Social-BIGAT(random) & Social-BIGAT(gates) \\
\midrule

    & eth &        1.11 +/- 0.04 &        1.09 +/- 0.04 &       \textbf{1.07 +/- 0.05} \\
    & hotel &        0.48 +/- 0.08 &        \textbf{0.45 +/- 0.05} &       0.47 +/- 0.12 \\
ADE(m) & univ &        0.79 +/- 0.04 &        0.73 +/- 0.04 &       \textbf{0.67 +/- 0.02} \\
    & zara1 &        0.56 +/- 0.02 &        0.53 +/- 0.02 &        \textbf{0.49 +/- 0.0} \\
    & zara2 &         0.5 +/- 0.03 &        0.48 +/- 0.02 &       \textbf{0.43 +/- 0.03} \\
    \hline
AVG   &  &        0.69 +/- 0.02 &        0.65 +/- 0.02 &       \textbf{0.62 +/- 0.04} \\
    \hline \hline

    & eth &        2.18 +/- 0.05 &        \textbf{2.16 +/- 0.07} &       2.16 +/- 0.09 \\
    & hotel &         0.9 +/- 0.16 &         \textbf{0.85 +/- 0.1} &       0.91 +/- 0.23 \\
FDE(m)    & univ &        1.57 +/- 0.09 &        1.46 +/- 0.08 &       \textbf{1.33 +/- 0.04} \\
    & zara1 &        1.14 +/- 0.04 &        1.11 +/- 0.03 &        \textbf{1.0 +/- 0.01} \\
    & zara2 &        0.98 +/- 0.05 &        0.94 +/- 0.04 &       \textbf{0.84 +/- 0.05} \\
    \hline
AVG  &  &        1.35 +/- 0.04 &         1.3 +/- 0.04 &       \textbf{1.25 +/- 0.08} \\
\bottomrule
\end{tabular}
\end{table*}

\begin{table*}
\caption{ADE and FDE in meters for Social-Ways on the three experiments(normal, random and gates)}
\label{fig:exp_ways}
\centering
\begin{tabular}{l|l|l|l|l}
\toprule
    &       & Social-Ways(normal) & Social-Ways(random) & Social-Ways(gates) \\
\midrule

    & eth &       1.08 +/- 0.02 &       1.06 +/- 0.04 &      \textbf{1.03 +/- 0.02} \\
    & hotel &        0.5 +/- 0.01 &       \textbf{0.47 +/- 0.05} &      0.53 +/- 0.06 \\
 ADE(m)   & univ &       0.76 +/- 0.03 &       0.68 +/- 0.05 &      \textbf{0.65 +/- 0.05} \\
    & zara1 &        0.5 +/- 0.01 &       0.49 +/- 0.01 &      \textbf{0.47 +/- 0.01} \\
    & zara2 &       0.45 +/- 0.01 &       0.43 +/- 0.01 &      \textbf{0.39 +/- 0.01} \\
    \hline
    AVG &  &       0.66 +/- 0.01 &       0.63 +/- 0.02 &      \textbf{0.62 +/- 0.01} \\
        \hline \hline

    & eth &       2.11 +/- 0.03 &       2.09 +/- 0.06 &      \textbf{2.08 +/- 0.05} \\
    & hotel &       0.94 +/- 0.04 &       \textbf{0.86 +/- 0.09} &      0.99 +/- 0.13 \\
 FDE(m)   & univ &       1.47 +/- 0.01 &        1.36 +/- 0.1 &      \textbf{1.29 +/- 0.07} \\
    & zara1 &       1.01 +/- 0.02 &       1.02 +/- 0.02 &      \textbf{0.97 +/- 0.01} \\
    & zara2 &       0.89 +/- 0.03 &       0.85 +/- 0.02 &      0.77 +/- 0.03 \\
    \hline
AVG  &  &       1.29 +/- 0.01 &       1.24 +/- 0.04 &      \textbf{1.22 +/- 0.03} \\
    \hline
\bottomrule
\end{tabular}
\end{table*}

\begin{figure}
\centering

\includegraphics[width=\linewidth]{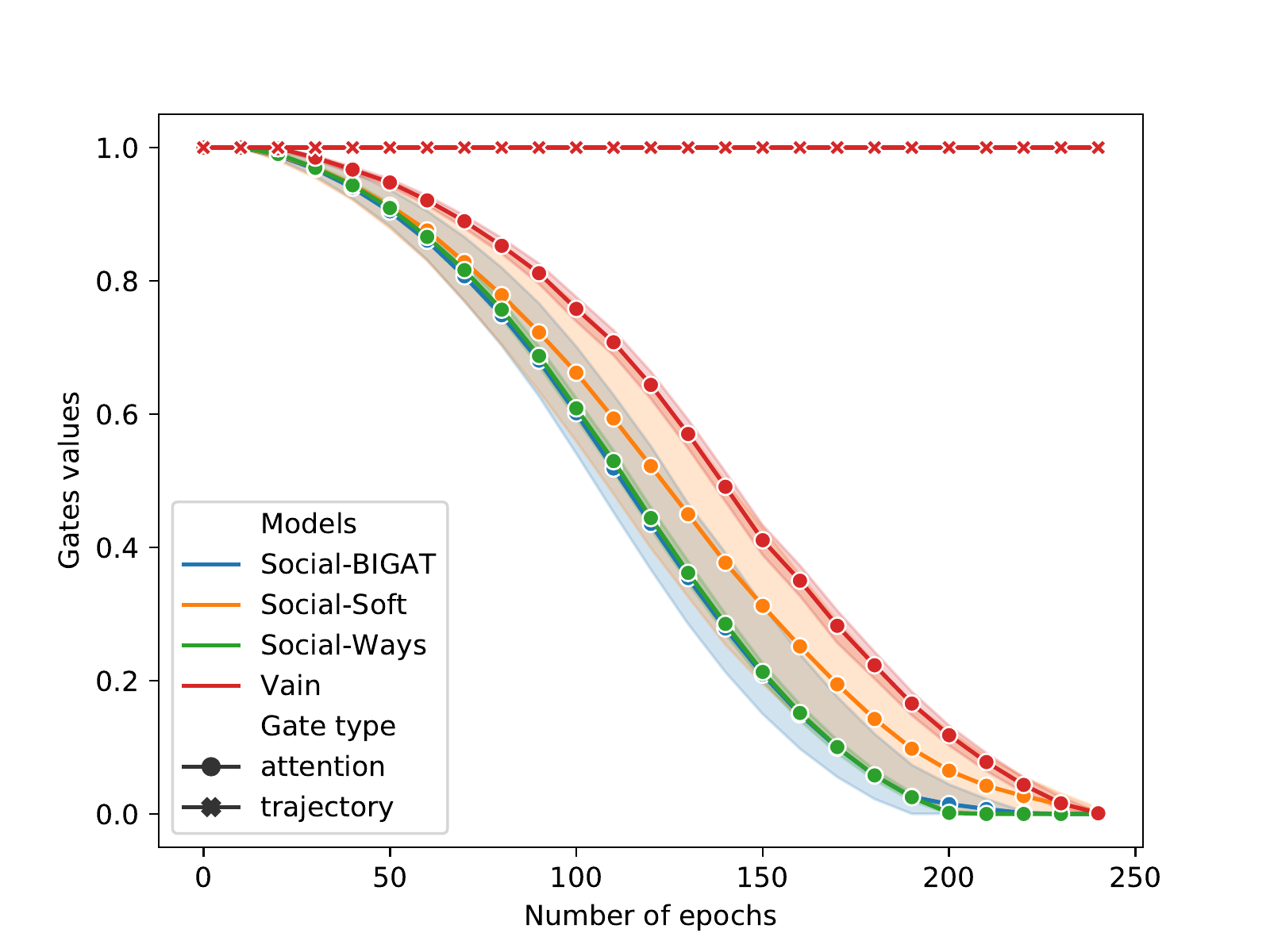}

\caption{ Evolution of gate values with the number of epochs when fine-tuning social models with a gating mechanism and $L_0$ penalty term. Results are shown when evaluating on scene \textsf{zara2}.}
\label{fig:evolution}
\end{figure}

\paragraph{Gate training} Finally, the models were first trained regularly and then fine-tuned by adding a gating mechanism on the \emph{social} and \emph{trajectory modules}, along with a $L_0$ term penalty as described in section~\ref{sec:gates}. Figure~\ref{fig:evolution} shows the evolution of the gates values during models fine-tuning. Since the same behavior is observed on the other scenes, we present it only on the scene \textsf{zara2}.
The attention gate is consistently set to zero whereas the trajectory gate always keeps its original value of one. Furthermore, to ensure that this evolution did not happen at the expense of the models prediction performance, the $ADE$ and $FDE$ of such fine-tuned models (green bars) are reported in  Table~\ref{fig:exp_bigat} for Social-Bigat under the column Social-Bigat(gates) , in Table~\ref{fig:exp_ways} for Social-Ways under the column Social-Ways(gates) and in  Table~\ref{fig:exp_vain} for Vain under the column Vain(gates). As for random training, the fine-tuned models all outperformed  their regularly trained counterparts. This time the improvement lies in the range of 0.05~m to 0.1~m for $ADE$ and from 0.07~m to 0.15~m for $FDE$. First of all, the improvement is not that large. When fine-tuning the models, the values of the gates vary and the models continue their convergence marginally. Since we apply fine-tuning for the same number of iterations as the pre-training step, this could explain the slight improvement observed in $ADE$ and $FDE$.
The models learned to ignore the output of the soft-attention mechanisms while keeping the information extracted from the pedestrian past trajectory. Moreover, it did not affect the convergence of the models which continued to improve slightly.

\begin{table*}
\caption{ADE and FDE in meters for Vain on the three experiments(normal, random and gates)}
\label{fig:exp_vain}
\centering
\begin{tabular}{l|l|l|l|l}
\toprule
    &       &   Vain(normal) &   Vain(random) &    Vain(gates) \\
\midrule

    & eth &  1.07 +/- 0.02 &  1.09 +/- 0.02 &  \textbf{1.03 +/- 0.01} \\
    & hotel &  0.48 +/- 0.02 &  0.46 +/- 0.05 &  \textbf{0.39 +/- 0.04} \\
 ADE(m)   & univ &   0.7 +/- 0.03 &   0.7 +/- 0.03 &  \textbf{0.68 +/- 0.02} \\
    & zara1 &  0.47 +/- 0.01 &  \textbf{0.46 +/- 0.01} &  0.47 +/- 0.01 \\
    & zara2 &  0.45 +/- 0.02 &  0.41 +/- 0.01 &  \textbf{0.39 +/- 0.01} \\
    \hline
 AVG    &  &  0.63 +/- 0.01 &  0.62 +/- 0.01 &  \textbf{0.59 +/- 0.01} \\
        \hline \hline

    & eth &   2.1 +/- 0.03 &  2.16 +/- 0.04 &  \textbf{2.05 +/- 0.03} \\
    & hotel &   0.9 +/- 0.05 &  0.86 +/- 0.09 &  \textbf{0.71 +/- 0.09} \\
 FDE(m)   & univ &  1.43 +/- 0.07 &  1.45 +/- 0.07 &  \textbf{1.35 +/- 0.06} \\
    & zara1 &  0.96 +/- 0.02 &  \textbf{0.96 +/- 0.01} &  0.97 +/- 0.02 \\
    & zara2 &  0.88 +/- 0.04 &  0.84 +/- 0.02 &  \textbf{0.79 +/- 0.04} \\
    \hline
 AVG    &  &  1.25 +/- 0.02 &  1.25 +/- 0.01 &  \textbf{1.17 +/- 0.03} \\
\bottomrule
\end{tabular}
\end{table*}

To summarize, training with random neighbours trajectories did not worsen the performances or convergences of the social models. When given the opportunities to shut down the \emph{attention} and \emph{trajectory modules}, the social models consistently shut down the \emph{social module}. This did not affect the training and the models even continued to slightly improve during the fine-tuning phase. These elements strongly indicate that the output of the soft-attention mechanisms is ignored by the models when predicting a trajectory. Therefore, we claim that the studied model's predictions are not made on the basis of social interactions.

\section{Conclusion}
We study whether soft-attention applied to social information has an impact on the trajectories predicted by four different social models. By means of two experiments, the output of the soft-attention mechanism is consistently ignored by the models. This implies those models do not use social information at prediction time. This is an important result as it contradicts the commonly accepted opposite statement. It highlights that ablative studies should be conducted more carefully when interpreting the impact of a component on a such complex task. 
We believe that the standard datasets and setup we used in this study should be set aside. On the one hand, the problem might be too easy for the model to actually have an interest in learning social interactions, as the good performances of the constant velocity model suggests. Predicting trajectory strips of approximately four seconds on those datasets mainly consists in predicting straight motions. On the other hand, modeling the impact of social interactions on the pedestrian trajectories is hard since the motion possibilities for pedestrian in low constraint environment such as the scenes in \textsf{ETH} and \textsf{UCY} are endless. It could be argued that new datasets, with more social interactions are now available and being used, so the results of our study  are not relevant. However the initial papers based their conclusions on those and inspired a plethora of \textit{socialX} methods following the same pattern. It is important to challenge the assumptions of those works and do so on the same datasets.
Future works should be aimed at predicting long-term trajectories on richer datasets such as the Intersection Drone Dataset \cite{bock2019ind}. It is also important to develop a new framework to demonstrate the need and ability of future models to learn social interactions. 



\bibliographystyle{IEEEtran}
\bibliography{neurips_2020}

\begin{thebibliography}{10}
\providecommand{\url}[1]{#1}
\csname url@samestyle\endcsname
\providecommand{\newblock}{\relax}
\providecommand{\bibinfo}[2]{#2}
\providecommand{\BIBentrySTDinterwordspacing}{\spaceskip=0pt\relax}
\providecommand{\BIBentryALTinterwordstretchfactor}{4}
\providecommand{\BIBentryALTinterwordspacing}{\spaceskip=\fontdimen2\font plus
\BIBentryALTinterwordstretchfactor\fontdimen3\font minus
  \fontdimen4\font\relax}
\providecommand{\BIBforeignlanguage}[2]{{%
\expandafter\ifx\csname l@#1\endcsname\relax
\typeout{** WARNING: IEEEtran.bst: No hyphenation pattern has been}%
\typeout{** loaded for the language `#1'. Using the pattern for}%
\typeout{** the default language instead.}%
\else
\language=\csname l@#1\endcsname
\fi
#2}}
\providecommand{\BIBdecl}{\relax}
\BIBdecl

\bibitem{alahi2016social}
A.~Alahi, K.~Goel, V.~Ramanathan, A.~Robicquet, L.~Fei-Fei, and S.~Savarese,
  ``Social lstm: Human trajectory prediction in crowded spaces,'' in
  \emph{Proceedings of the IEEE conference on computer vision and pattern
  recognition}, 2016, pp. 961--971.

\bibitem{gupta2018social}
A.~Gupta, J.~Johnson, L.~Fei-Fei, S.~Savarese, and A.~Alahi, ``Social gan:
  Socially acceptable trajectories with generative adversarial networks,'' in
  \emph{Proceedings of the IEEE Conference on Computer Vision and Pattern
  Recognition}, 2018, pp. 2255--2264.

\bibitem{sadeghian2019sophie}
A.~Sadeghian, V.~Kosaraju, A.~Sadeghian, N.~Hirose, H.~Rezatofighi, and
  S.~Savarese, ``Sophie: An attentive gan for predicting paths compliant to
  social and physical constraints,'' in \emph{Proceedings of the IEEE
  Conference on Computer Vision and Pattern Recognition}, 2019, pp. 1349--1358.

\bibitem{xu2015show}
K.~Xu, J.~Ba, R.~Kiros, K.~Cho, A.~Courville, R.~Salakhudinov, R.~Zemel, and
  Y.~Bengio, ``Show, attend and tell: Neural image caption generation with
  visual attention,'' in \emph{International conference on machine learning},
  2015, pp. 2048--2057.

\bibitem{hoshen2017vain}
Y.~Hoshen, ``Vain: Attentional multi-agent predictive modeling,'' in
  \emph{Advances in Neural Information Processing Systems}, 2017, pp.
  2701--2711.

\bibitem{amirian2019social}
J.~Amirian, J.-B. Hayet, and J.~Pettr{\'e}, ``Social ways: Learning multi-modal
  distributions of pedestrian trajectories with gans,'' in \emph{Proceedings of
  the IEEE Conference on Computer Vision and Pattern Recognition Workshops},
  2019, pp. 0--0.

\bibitem{robicquet2016learning}
A.~Robicquet, A.~Sadeghian, A.~Alahi, and S.~Savarese, ``Learning social
  etiquette: Human trajectory understanding in crowded scenes,'' in
  \emph{European conference on computer vision}.\hskip 1em plus 0.5em minus
  0.4em\relax Springer, 2016, pp. 549--565.

\bibitem{nikhil2018convolutional}
N.~Nikhil and B.~Tran~Morris, ``Convolutional neural network for trajectory
  prediction,'' in \emph{Proceedings of the European Conference on Computer
  Vision (ECCV)}, 2018, pp. 0--0.

\bibitem{becker2018red}
S.~Becker, R.~Hug, W.~Hubner, and M.~Arens, ``Red: A simple but effective
  baseline predictor for the trajnet benchmark,'' in \emph{Proceedings of the
  European Conference on Computer Vision (ECCV)}, 2018, pp. 0--0.

\bibitem{scholler2019constant}
C.~Sch{\"o}ller, V.~Aravantinos, F.~Lay, and A.~Knoll, ``What the constant
  velocity model can teach us about pedestrian motion prediction,'' in
  \emph{arXiv: 1903.07933}, 2019.

\bibitem{kosaraju2019social}
V.~Kosaraju, A.~Sadeghian, R.~Mart{\'\i}n-Mart{\'\i}n, I.~Reid, H.~Rezatofighi,
  and S.~Savarese, ``Social-bigat: Multimodal trajectory forecasting using
  bicycle-gan and graph attention networks,'' in \emph{Advances in Neural
  Information Processing Systems}, 2019, pp. 137--146.

\bibitem{voita2019analyzing}
E.~Voita, D.~Talbot, F.~Moiseev, R.~Sennrich, and I.~Titov, ``Analyzing
  multi-head self-attention: Specialized heads do the heavy lifting, the rest
  can be pruned,'' \emph{arXiv preprint arXiv:1905.09418}, 2019.

\bibitem{louizos2017learning}
C.~Louizos, M.~Welling, and D.~P. Kingma, ``Learning sparse neural networks
  through $ l\_0 $ regularization,'' \emph{arXiv preprint arXiv:1712.01312},
  2017.

\bibitem{lefevre2014survey}
S.~Lef{\`e}vre, D.~Vasquez, and C.~Laugier, ``A survey on motion prediction and
  risk assessment for intelligent vehicles,'' \emph{ROBOMECH journal}, vol.~1,
  no.~1, p.~1, 2014.

\bibitem{ammoun2009real}
S.~Ammoun and F.~Nashashibi, ``Real time trajectory prediction for collision
  risk estimation between vehicles,'' in \emph{2009 IEEE 5th International
  Conference on Intelligent Computer Communication and Processing}.\hskip 1em
  plus 0.5em minus 0.4em\relax IEEE, 2009, pp. 417--422.

\bibitem{welch1995introduction}
G.~Welch, G.~Bishop \emph{et~al.}, ``An introduction to the kalman filter,''
  1995.

\bibitem{lin2000vehicle}
C.-F. Lin, A.~G. Ulsoy, and D.~J. LeBlanc, ``Vehicle dynamics and external
  disturbance estimation for vehicle path prediction,'' \emph{IEEE Transactions
  on Control Systems Technology}, vol.~8, no.~3, pp. 508--518, 2000.

\bibitem{morris2011trajectory}
B.~T. Morris and M.~M. Trivedi, ``Trajectory learning for activity
  understanding: Unsupervised, multilevel, and long-term adaptive approach,''
  \emph{IEEE transactions on pattern analysis and machine intelligence},
  vol.~33, no.~11, pp. 2287--2301, 2011.

\bibitem{helbing1995social}
D.~Helbing and P.~Molnar, ``Social force model for pedestrian dynamics,''
  \emph{Physical review E}, vol.~51, no.~5, p. 4282, 1995.

\bibitem{ballan2016knowledge}
L.~Ballan, F.~Castaldo, A.~Alahi, F.~Palmieri, and S.~Savarese, ``Knowledge
  transfer for scene-specific motion prediction,'' in \emph{European Conference
  on Computer Vision}.\hskip 1em plus 0.5em minus 0.4em\relax Springer, 2016,
  pp. 697--713.

\bibitem{vemula2017modeling}
A.~Vemula, K.~Muelling, and J.~Oh, ``Modeling cooperative navigation in dense
  human crowds,'' in \emph{2017 IEEE International Conference on Robotics and
  Automation (ICRA)}.\hskip 1em plus 0.5em minus 0.4em\relax IEEE, 2017, pp.
  1685--1692.

\bibitem{simonyan2014very}
K.~Simonyan and A.~Zisserman, ``Very deep convolutional networks for
  large-scale image recognition,'' \emph{arXiv preprint arXiv:1409.1556}, 2014.

\bibitem{sadeghian2018car}
A.~Sadeghian, F.~Legros, M.~Voisin, R.~Vesel, A.~Alahi, and S.~Savarese,
  ``Car-net: Clairvoyant attentive recurrent network,'' in \emph{Proceedings of
  the European Conference on Computer Vision (ECCV)}, 2018, pp. 151--167.

\bibitem{hug2017reliability}
R.~Hug, S.~Becker, W.~H{\"u}bner, and M.~Arens, ``On the reliability of
  lstm-mdl models for pedestrian trajectory prediction,'' in
  \emph{International Workshop on Representations, Analysis and Recognition of
  Shape and Motion From Imaging Data}.\hskip 1em plus 0.5em minus 0.4em\relax
  Springer, 2017, pp. 20--34.

\bibitem{sohn2015learning}
K.~Sohn, H.~Lee, and X.~Yan, ``Learning structured output representation using
  deep conditional generative models,'' in \emph{Advances in neural information
  processing systems}, 2015, pp. 3483--3491.

\bibitem{lee2017desire}
N.~Lee, W.~Choi, P.~Vernaza, C.~B. Choy, P.~H. Torr, and M.~Chandraker,
  ``Desire: Distant future prediction in dynamic scenes with interacting
  agents,'' in \emph{Proceedings of the IEEE Conference on Computer Vision and
  Pattern Recognition}, 2017, pp. 336--345.

\bibitem{goodfellow2014generative}
I.~Goodfellow, J.~Pouget-Abadie, M.~Mirza, B.~Xu, D.~Warde-Farley, S.~Ozair,
  A.~Courville, and Y.~Bengio, ``Generative adversarial nets,'' in
  \emph{Advances in neural information processing systems}, 2014, pp.
  2672--2680.

\bibitem{varshneya2017human}
D.~Varshneya and G.~Srinivasaraghavan, ``Human trajectory prediction using
  spatially aware deep attention models,'' \emph{arXiv preprint
  arXiv:1705.09436}, 2017.

\bibitem{deo2018convolutional}
N.~Deo and M.~M. Trivedi, ``Convolutional social pooling for vehicle trajectory
  prediction,'' in \emph{Proceedings of the IEEE Conference on Computer Vision
  and Pattern Recognition Workshops}, 2018, pp. 1468--1476.

\bibitem{huang2019stgat}
Y.~Huang, H.~Bi, Z.~Li, T.~Mao, and Z.~Wang, ``Stgat: Modeling spatial-temporal
  interactions for human trajectory prediction,'' in \emph{Proceedings of the
  IEEE International Conference on Computer Vision}, 2019, pp. 6272--6281.

\bibitem{ma2019trafficpredict}
Y.~Ma, X.~Zhu, S.~Zhang, R.~Yang, W.~Wang, and D.~Manocha, ``Trafficpredict:
  Trajectory prediction for heterogeneous traffic-agents,'' in
  \emph{Proceedings of the AAAI Conference on Artificial Intelligence},
  vol.~33, 2019, pp. 6120--6127.

\bibitem{pellegrini2009you}
S.~Pellegrini, A.~Ess, K.~Schindler, and L.~Van~Gool, ``You'll never walk
  alone: Modeling social behavior for multi-target tracking,'' in \emph{2009
  IEEE 12th International Conference on Computer Vision}.\hskip 1em plus 0.5em
  minus 0.4em\relax IEEE, 2009, pp. 261--268.

\bibitem{xu2018cidnn}
Y.~Xu, Z.~Piao, and S.~Gao, ``Encoding crowd interaction with deep neural
  network for pedestrian trajectory prediction,'' in \emph{2018 IEEE Conference
  on Computer Vision and Pattern Recognition (CVPR)}, 2018.

\bibitem{velivckovic2017graph}
P.~Veli{\v{c}}kovi{\'c}, G.~Cucurull, A.~Casanova, A.~Romero, P.~Lio, and
  Y.~Bengio, ``Graph attention networks,'' \emph{arXiv preprint
  arXiv:1710.10903}, 2017.

\bibitem{kingma2013auto}
D.~P. Kingma and M.~Welling, ``Auto-encoding variational bayes,'' \emph{arXiv
  preprint arXiv:1312.6114}, 2013.

\bibitem{lerner2007crowds}
A.~Lerner, Y.~Chrysanthou, and D.~Lischinski, ``Crowds by example,'' in
  \emph{Computer graphics forum}, vol.~26, no.~3.\hskip 1em plus 0.5em minus
  0.4em\relax Wiley Online Library, 2007, pp. 655--664.

\bibitem{bock2019ind}
J.~Bock, R.~Krajewski, T.~Moers, S.~Runde, L.~Vater, and L.~Eckstein, ``The ind
  dataset: A drone dataset of naturalistic road user trajectories at german
  intersections,'' \emph{arXiv preprint arXiv:1911.07602}, 2019.

\end{thebibliography}

\end{document}